\documentclass[letterpaper]{article} 
\usepackage{aaai2026}  
\usepackage{times}  
\usepackage{helvet}  
\usepackage{courier}  
\usepackage[hyphens]{url}  
\usepackage{graphicx} 
\usepackage{natbib}  
\usepackage{caption} 
\usepackage{latexsym}
\usepackage{amssymb}
\usepackage{amsmath}
\usepackage{amsthm}
\usepackage{booktabs}
\usepackage{enumitem}
\usepackage{graphicx}
\usepackage{color}
\usepackage{dsfont}
\usepackage{multirow}
\usepackage{bm}
\newcommand{\figref}[1]{Figure \ref{#1}}
\newcommand{\tabref}[1]{Table \ref{#1}}

\newcommand{\equref}[1]{Equation (\ref{#1})}

\title{Global-Lens Transformers: Adaptive Token Mixing for Dynamic Link Prediction}
\author {
    Tao Zou\textsuperscript{\rm 1},
    Chengfeng Wu\textsuperscript{\rm 2},
    Tianxi Liao\textsuperscript{\rm 1},
    Junchen Ye\textsuperscript{\rm 3}\thanks{Corresponding author.},
    Bowen Du\textsuperscript{\rm 3,4}
}

\affiliations {
    \textsuperscript{\rm 1}State Key Laboratory of Complex \& Critical Software Environment, Beihang University, Beijing, China\\
    \textsuperscript{\rm 2}Shenzhen Key Laboratory of Ubiquitous Data Enabling, Tsinghua Shenzhen International Graduate School, Shenzhen, China\\
    \textsuperscript{\rm 3}School of Transportation Science and Engineering, Beihang University, Beijing, China\\
    \textsuperscript{\rm 4}Zhongguancun Laboratory, Beijing, China\\
    zoutao@buaa.edu.cn, wucf25@mails.tsinghua.edu.cn, \{tx\_liao, junchenye, dubowen\}@buaa.edu.cn
}
\begin{document}

\maketitle


\begin{abstract}
Dynamic graph learning plays a pivotal role in modeling evolving relationships over time, especially for temporal link prediction tasks in domains such as traffic systems, social networks, and recommendation platforms. While Transformer-based models have demonstrated strong performance by capturing long-range temporal dependencies, their reliance on self-attention results in quadratic complexity with respect to sequence length, limiting scalability on high-frequency or large-scale graphs.
In this work, we revisit the necessity of self-attention in dynamic graph modeling. Inspired by recent findings that attribute the success of Transformers more to their architectural design than attention itself, we propose \textbf{GLFormer}, a novel attention-free Transformer-style framework for dynamic graphs. GLFormer introduces an \textbf{adaptive token mixer} that performs context-aware local aggregation based on interaction order and time intervals. To capture long-term dependencies, we further design a \textbf{hierarchical aggregation module} that expands the temporal receptive field by stacking local token mixers across layers.
Experiments on six widely used dynamic graph benchmarks show that GLFormer achieves competitive or superior performance, which reveals that attention-free architectures can match or surpass Transformer baselines in dynamic graph settings with significantly improved efficiency.
\end{abstract}


\section{Introduction}
\label{section-1}







Graph-structured data is prevalent in numerous real-world applications, such as traffic systems~\cite{DBLP:conf/aaai/JiW0JZ22, DBLP:journals/tkde/JinLFSHZZ24}, social networks~\cite{DBLP:conf/www/ZhouLDJ023, DBLP:journals/tweb/JainKS23}, and recommendation systems~\cite{DBLP:conf/kdd/Jiang0H23, DBLP:journals/tois/Zhang0YD023}. In these graphs, nodes interact over time, forming temporally ordered sequences of events that govern the evolution of the network. Effectively modeling such dynamics is critical for tasks like future link prediction, recommendation, and fraud detection.

A growing line of research has adopted Transformer-based architectures for dynamic graph learning due to their strong capacity for capturing long-range dependencies across sequences of interactions~\cite{DBLP:journals/corr/abs-2105-07944, DBLP:conf/nips/0004S0L23, DBLP:conf/nips/VaswaniSPUJGKP17}. By attending over time-stamped neighbors, these models can learn expressive representations for evolving nodes. Existing approaches commonly follow a structured pipeline: capture the structural information from temporal graphs by using time-aware random walks~\cite{DBLP:conf/bigdataconf/NguyenLRAKK18, yu2024genti}, and memory networks~\cite{DBLP:conf/kdd/KumarZL19, DBLP:journals/corr/abs-2006-10637}; then apply Transformers \cite{DBLP:conf/nips/0004S0L23, pan2025light} to learn temporal dependencies across the historical interaction sequences. 

\begin{figure}[!htbp]
    \centering
    \includegraphics[width=1.00\columnwidth]{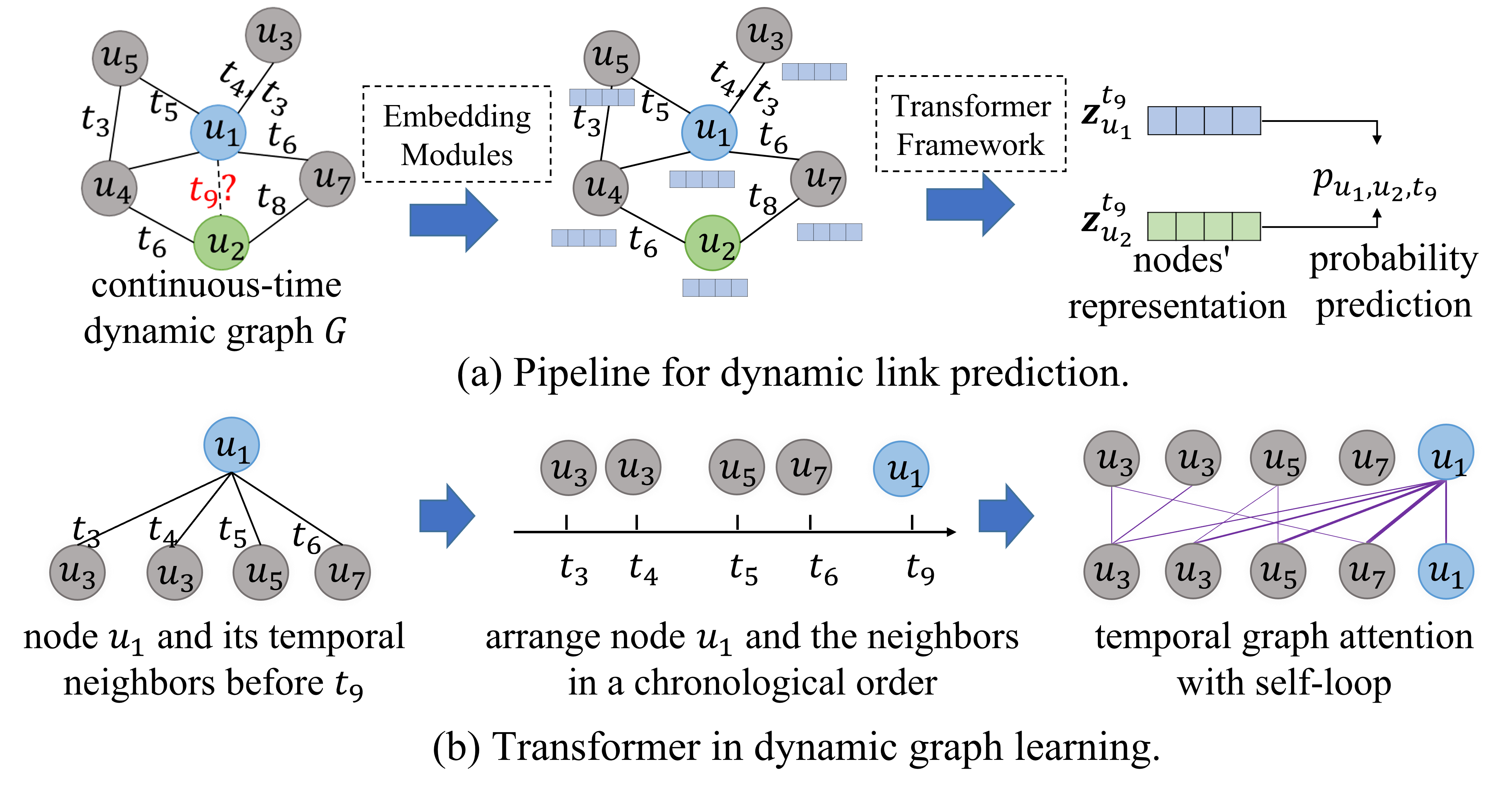}
    \caption{Illustration of a dynamic graph $G$ that evolves from $t_0$ to $t_8$ in (a). We aim to predict whether $u_2$ will interact with $u_1$ at timestamp $t_9$. (b) To capture the temporal dependencies among neighbors, existing works use self-attention mechanisms to learn these correlations.}
    \label{fig:dynamic graph}
\end{figure}

Figure~\ref{fig:dynamic graph} illustrates a dynamic graph link prediction task and the common modeling practice using temporal attention. However, \textbf{the self-attention mechanism scales quadratically with sequence length}, posing computational challenges for large-scale or high-frequency graphs. Moreover, attention indiscriminately aggregates all pairwise interactions, which can amplify noise and degrade generalization.

\begin{figure}[!htbp]
    \centering
    \includegraphics[width=1.00\columnwidth]{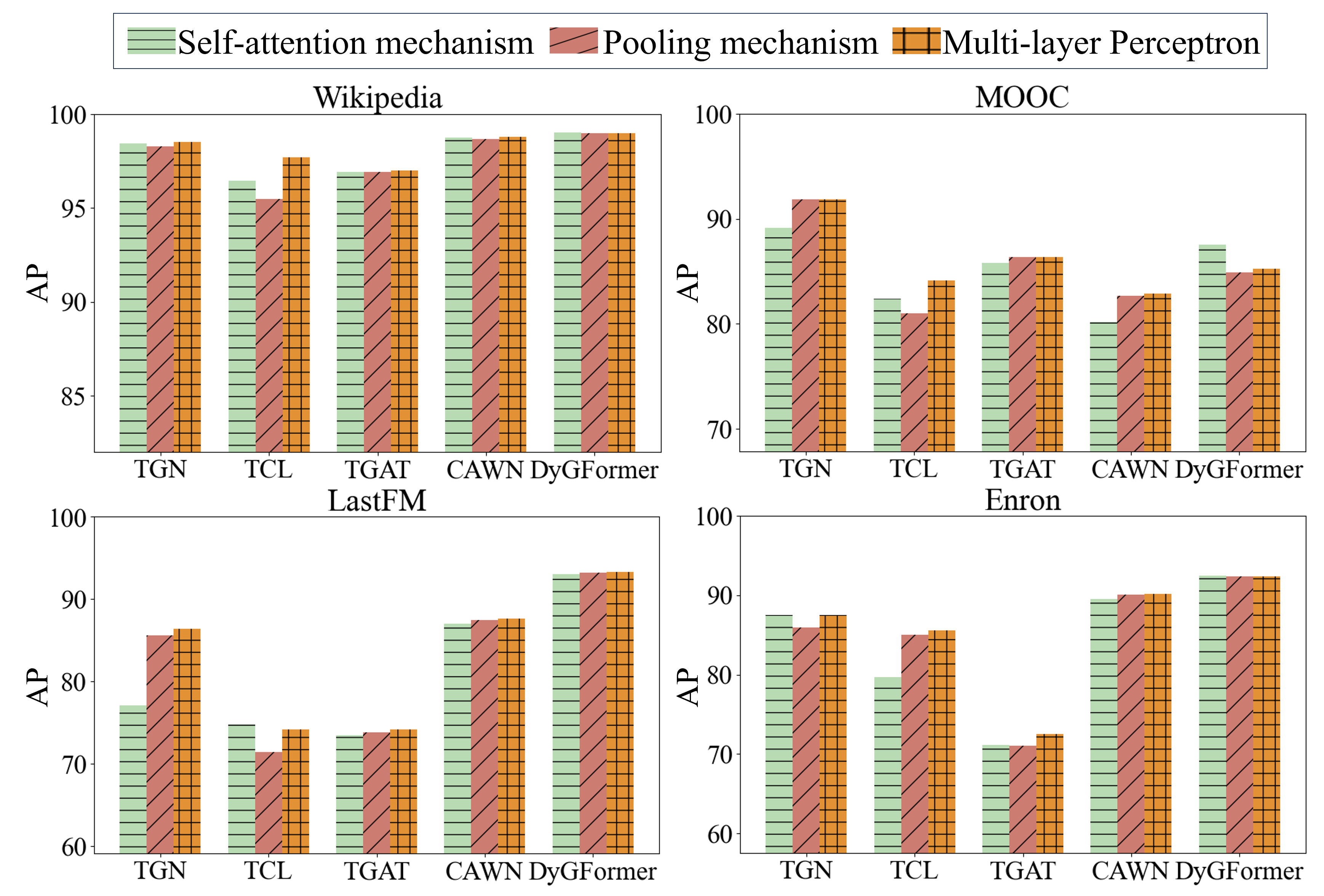}
    \caption{We show the average precision results for dynamic link prediction on four datasets with three types of token aggregation mechanisms.}
    \label{fig:motivation}
\end{figure}

Interestingly, recent studies in computer vision and sequence modeling~\cite{DBLP:conf/cvpr/YuLZSZWFY22, DBLP:journals/pami/YuSZLZFYW24} have shown that the success of Transformer architectures may stem more from their overarching structure than the self-attention mechanism itself. This observation raises an important question: \textit{Is it possible to design simpler, attention-free architectures for dynamic graphs that still retain strong representational power?} To explore this, we conduct controlled experiments by replacing self-attention with pooling and MLP modules across five Transformer baselines. As shown in Figure~\ref{fig:motivation}, these attention-free variants often match the original Transformer’s performance across four datasets. This suggests that efficient architectures for dynamic graph learning can be achieved without the computational overhead of attention.

Motivated by these insights, we propose \textbf{GLFormer}, a lightweight Transformer-style model for dynamic graphs. Instead of self-attention, GLFormer adopts an \textbf{adaptive token mixer} that aggregates tokens based on their temporal order and time intervals, allowing the model to capture local temporal context in a more efficient and structured manner. To further enhance representation power, we introduce a \textbf{hierarchical aggregation module} that expands the temporal receptive field by stacking token mixers across layers. This combination effectively models both short- and long-range dependencies with significantly lower computational cost. The contributions are as follows.

\begin{itemize}
    \item We propose \textbf{GLFormer}, a lightweight Transformer-style framework for dynamic graph learning. It replaces self-attention with an \textbf{adaptive token mixer} that leverages temporal order and interaction intervals, improving efficiency and robustness on evolving neighbor sequences.

    \item We introduce a \textbf{hierarchical aggregation mechanism} that progressively expands the temporal receptive field, enabling the model to capture long-range dependencies without incurring the cost of global attention.

    \item We conduct comprehensive experiments on six dynamic graph datasets. GLFormer consistently outperforms existing methods (e.g., TGAT, DyGFormer) in link prediction accuracy, while offering significantly faster inference, demonstrating its practicality and effectiveness.
\end{itemize}

\section{Related Work}
\label{section-5}
\subsection{Dynamic Graph Learning}

Dynamic graph learning aims to model the evolving patterns of nodes and their interactions by learning temporal latent representations \cite{DBLP:journals/csur/BarrosMVZ23, xu2025fast}. A core task in this field is \textit{dynamic link prediction} \cite{DBLP:conf/kdd/0003PWCY025, xu2025unidyg}, which estimates the probability of future links at a given time. Existing methods can be broadly categorized into two types: discrete-time and continuous-time approaches. \textit{Discrete-time} methods \cite{xu2024timesgn, chen2024dtformer} model dynamic graphs as sequences of snapshots, each summarizing interactions within a fixed interval. They typically use static graph encoders for structure and sequential models for temporal dynamics. However, this coarse partitioning overlooks the exact order of events, limiting their ability to capture fine-grained temporal patterns. In contrast, \textit{continuous-time} methods model the graph as a stream of timestamped interaction events. Approaches such as temporal random walks \cite{DBLP:conf/kdd/YuCAZCW18, yu2024genti} and memory-based representations \cite{DBLP:conf/kdd/KumarZL19, ji2024memmap} have been proposed to encode evolving topologies. These models often incorporate attention mechanisms or temporal neural networks \cite{DBLP:conf/nips/0004S0L23, pan2025light, DBLP:conf/kdd/0003MY024, DBLP:conf/kdd/0003PYS024, ding2024dygmamba} to capture long-term dependencies at the event level.

Despite their effectiveness, most existing models rely heavily on standard Transformer architectures, which incur high computational overhead. In this paper, we propose an efficient Transformer-based framework that integrates a low-cost adaptive aggregation mechanism for dynamic link prediction.



\subsection{Transformers in Various Fields}
Transformers have shown superiority in multiple fields, such as computer vision \cite{DBLP:journals/tmlr/SteinerKZWUB22, DBLP:conf/nips/RaoZT0LL22}, time series \cite{DBLP:conf/aaai/WangCC24, DBLP:conf/iclr/WangZWGD024}, and graphs \cite{DBLP:conf/nips/KreuzerBHLT21, DBLP:conf/icml/ChenOB22} beyond natural language processing. Vision Transformers (ViTs) \cite{DBLP:conf/iclr/DosovitskiyB0WZ21} treat images as sequences of patches for tasks like classification and detection. Similarly, in time series, Transformers capture temporal dependencies for forecasting \cite{DBLP:conf/aaai/WangCC24}, and Graph Transformers integrate structural information using methods like Laplacian eigenvectors or relative position encoding \cite{DBLP:conf/nips/KreuzerBHLT21}.
Recently, there has been a trend of replacing complicated self-attention modules with simpler architectures. Examples include Metaformer \cite{DBLP:conf/cvpr/YuLZSZWFY22} in computer vision and simplified attention designs like those in Informer \cite{DBLP:conf/aaai/ZhouZPZLXZ21} and using pyramid attention \cite{DBLP:conf/iclr/LiuYLLLLD22} for time series.

Although the above methods have demonstrated the power of simple architectures in various applications, they are not specialized for dynamic graph learning. In this paper, we aim to design a succinct architecture specifically for link predictions.


\section{Preliminaries}
\label{section-2}
\subsection{Problem Formulation}
\textbf{Definition 1. Dynamic Graph.} Given a set of nodes $\mathcal{V}$, the dynamic graph, denoted as $\mathcal{G}_{t_k}$, is a sequence of timestamped interactions: $\mathcal{G}_{t_k} = \{(u_1, v_1, t_1), \ldots, (u_k, v_k, t_k)\}$, where 
$u_i, v_i \in \mathcal{V}$ and $0 \leq t_1 \leq \ldots \leq t_k$. For an interaction $e_i=(u_i, v_i, t_i)$ that appears at timestamp $t_i$, $u_i$ and $v_i$ represent the source node and destination node respectively. Each node $u_i$ in the graph is equipped with a node feature $\bm{x}_{u_i} \in \mathbb{R}^{d_N}$ and each interaction $e_i$ is associated with an edge feature $\bm{x}_{e_i} \in \mathbb{R}^{d_E}$, where $\mathbb{R}^{d_N}, \mathbb{R}^{d_E}$ denote the dimensions of node and edge features.

\textbf{Definition 2. Dynamic Link Prediction.} Given the interaction $e=(u, v, t)$, dynamic graph learning aims to design a model for learning the temporal representations $\bm{Z}_{u}^t, \bm{Z}_{v}^t$ for nodes $u$ and $v$ based on all past interactions $\{(u_i, v_i, t_i) \in \mathcal{G} | t_i < t\}$. The ultimate goal of dynamic link prediction is to forecast whether the interaction $e$ will take place at time $t$, that is, to determine if $(u, v, t) \in \mathcal{G}$.

\subsection{Transformer Architecture}
In dynamic graph learning, the Transformer architecture~\cite{DBLP:conf/nips/VaswaniSPUJGKP17} is widely used to encode temporal neighborhood sequences. For instance, TGN~\cite{DBLP:journals/corr/abs-2006-10637} and TGAT~\cite{DBLP:conf/iclr/XuRKKA20} stack $K$ attention layers to aggregate information from $K$-hop neighbors, while DyGFormer~\cite{DBLP:conf/nips/0004S0L23} leverages the Transformer to capture long-term temporal dependencies from first-order neighborhoods. We here present a generic formulation of the Transformer-based embedding module for dynamic graphs.

Let $\bm{H}=[\bm{h}_1,\ldots,\bm{h}_N]\in\mathbb{R}^{N\times d}$ denote the sequence of node $u_i$'s neighbor embeddings. Single-head scaled dot-product attention computes
\begin{align}
\label{eq:self-attn}
\bm{Q} = \bm{H}\bm{W}_Q,\qquad
&\bm{K} = \bm{H}\bm{W}_K,\qquad
\bm{V} = \bm{H}\bm{W}_V, \\
\label{eq:softmax-weights}
\alpha_{i,j} &=
\frac{\exp\!\left(\frac{\bm{q}_i \bm{k}_j^\top}{\sqrt{d_k}}\right)}
{\sum_{m=1}^{N}\exp\!\left(\frac{\bm{q}_i \bm{k}_m^\top}{\sqrt{d_k}}\right)}, \\
\label{self-attn}
\bm{h}^{\mathrm{SA}}_{i} &= \sum_{j=1}^{N} \alpha_{i,j}\,\bm{v}_j,
\end{align}
where $\bm{W}_Q,\bm{W}_K\!\in\!\mathbb{R}^{d\times d_k}$ and $\bm{W}_V\!\in\!\mathbb{R}^{d\times d_v}$ are learnable parameters; $\bm{q}_i,\bm{k}_j\!\in\!\mathbb{R}^{d_k}$ and $\bm{v}_j\!\in\!\mathbb{R}^{d_v}$. Stacking $\bm{h}^{\mathrm{SA}}_i$ over $i$ yields $\bm{H}_{\mathrm{SA}}\in\mathbb{R}^{N\times d_v}$. In compact matrix form,
\begin{equation}
    \mathrm{Attn}(\bm{H})=\mathrm{softmax}\!\big(\bm{Q}\bm{K}^\top/\sqrt{d_k}\big)\,\bm{V},
\end{equation}
where the softmax is applied row-wise.

We then project the attention output back to the model dimension and apply a residual connection, followed by a feed-forward network (FFN) with layer normalization (LN):
\begin{align}
\label{eq:residual}
\widehat{\bm{H}} &= \bm{H} + \bm{H}_{\mathrm{SA}}\bm{W}_O, \qquad \bm{W}_O\in\mathbb{R}^{d_v\times d},\\
\label{eq:ffn}
\bm{H}_{\mathrm{TA}} &= \widehat{\bm{H}} + \mathrm{FFN}\!\big(\mathrm{LN}(\widehat{\bm{H}})\big).
\end{align}
When $d_v=d$, \equref{eq:residual} reduces to $\widehat{\bm{H}}=\bm{H}+\bm{H}_{\mathrm{SA}}$. The multi-head extension uses head-specific projections $\{\bm{W}_Q^{(h)},\bm{W}_K^{(h)},\bm{W}_V^{(h)}\}_{h=1}^H$, concatenates head outputs, and applies a shared output projection $\bm{W}_O$.

\section{Methodology}
\label{section-3}
In this section, we introduce GLFormer, an efficient Transformer framework designed for learning temporal dependencies in dynamic graphs. Specifically, given a node's sequence of neighbors, we first obtain the representations of each neighbor through an embedding layer, which is derived from existing dynamic graph learning methods. Subsequently, we employ a unified Transformer framework to capture the temporal dependencies among these neighbors. This framework accepts a sequence of token representations as input and incorporates adaptive token-wise aggregation alongside a standard channel-wise feedforward network. To effectively model long-term relationships with neighbors, we propose a hierarchical aggregation mechanism. Ultimately, GLFormer outputs the temporal representations of the sequence, enabling accurate link prediction.
\begin{figure}[!htbp]
    \centering
\includegraphics[width=0.8\columnwidth]{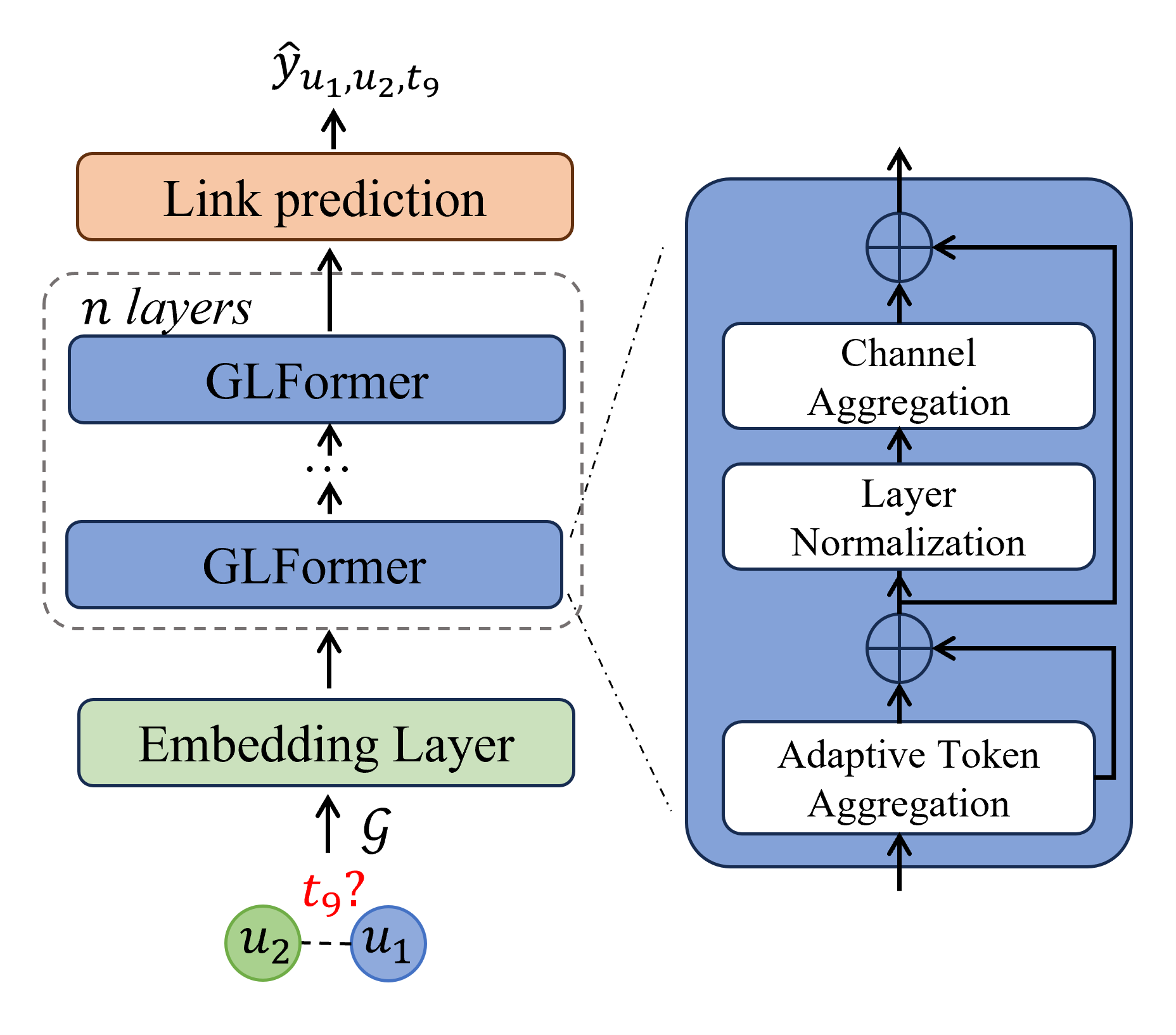}
    \caption{Framework of GLFormer.}
    \label{fig:framework}
\end{figure}
\subsection{GLFormer}
Previous research has successfully applied Transformer architectures to model long-range dependencies in various fields, such as computer vision \cite{DBLP:conf/iclr/DosovitskiyB0WZ21, DBLP:journals/tmlr/SteinerKZWUB22, DBLP:conf/nips/RaoZT0LL22}, natural language processing \cite{DBLP:conf/iclr/PatelLRCRC23, DBLP:conf/naacl/DevlinCLT19}, and time series \cite{DBLP:conf/aaai/WangCC24, DBLP:conf/iclr/WangZWGD024, DBLP:conf/nips/NiYLLL23}. Recent works \cite{DBLP:conf/cvpr/YuLZSZWFY22, DBLP:journals/pami/YuSZLZFYW24} have observed that the impressive performance often stems from the general architecture rather than the self-attention mechanism itself. Consequently, they replaced the self-attention module with relatively low-complexity components. Inspired by these findings, we investigate the utility of the self-attention mechanism in dynamic graph learning and find a similar tendency, as illustrated in \figref{fig:motivation}. Therefore, in this work, we propose a simplified yet efficient Transformer architecture for dynamic graphs.

Firstly, given an interaction $e=(u, v, t)$, we derive the initial embeddings of nodes from the embedding module of existing temporal graph learning methods. These methods employ different types of embedding processes, such as temporal random walks, memory-based updating mechanisms, and Multi-Layer Perceptrons (MLPs). The process can be generalized as follows,
\begin{equation}
    \bm{X}_{u} = f_{E}\left(\mathcal{G}_{t^{-}}, u\right),
\end{equation}
where $f_{E}(\cdot)$ denotes the embedding function and $\bm{X}_{u}$ is the initial embedding of node $u$.\footnote{Note that the initial embeddings should be $\bm{X}_{u}^{t^{-}}$, we ignore the timestamp $t^{-}$ in the following for simplification.}


After obtaining the embeddings of the nodes, we utilize GLFormer to capture the temporal relationships across the entire graph. The input $\bm{I}_u = [\bm{X}_{u_1}, \ldots, \bm{X}_{u_N}] \in \mathbb{R}^{N \times d}$ represents the embeddings of node $u$'s neighbors, where $N$ indicates the number of neighbors and $d$ is the embedding dimension. GLFormer comprises two sub-blocks: a token mixer that aggregates information from the tokens and a channel mixer that learns the relationships between these tokens, with each block incorporating a residual network to integrate the inputs seamlessly.

\textbf{Adaptive Token Aggregation Module.} Existing methods utilize the standard attention mechanism to aggregate temporal information among tokens, as illustrated in \equref{self-attn}. This can be interpreted as a spatial smoothing technique, corroborated by findings in several studies \cite{DBLP:conf/iccv/LiuL00W0LG21, DBLP:conf/iclr/ParkK22}. In dynamic graph learning, recent neighbors inherently provide the most relevant and informative interaction patterns. To smooth the information, we propose an adaptive local aggregation mechanism that considers both the timing and order of interactions. Specifically, for each neighbor $u_i$ interacting at $t_i$, we first select the most recent $M$ neighbors up to $t_i$ for aggregation. We employ learnable parameters $\bm{W}$ to capture the importance of order, while the significance of timing $\bm{\theta}^i$ is measured by applying a softmax function to the relative time differences. We fuse these two critical factors with a learnable parameter $\beta$, as shown below,
\begin{align}
    \bm{H}_{i,:} & = \sum_{p=0}^{M-1} \alpha_{p}^{i}\bm{I}_{i-p, :}, \\
    \alpha_{p}^{i} &= \beta \bm{w}_{p} + (1-\beta)\theta_{p}^{i}, \\
    \theta_{p}^{i} &= \frac{\exp\!\big(-\big(t_i - t_{i-p}\big)\big)}{\sum_{q=0}^{M-1}\exp\!\big(-\big(t_i - t_{i-q}\big)\big)},
\end{align}
where the neighbor $u_i$'s representation $\bm{H}_{i,:}$ is obtained via a weighted average of neighboring embeddings $\bm{I}_{i-p,:}$, with scores $\alpha_{p}^{i}$ determined by both local learnable parameters $\bm{w}_{p}$ and a time-based factor $\theta_{p}^{i}$. For positions near the beginning of the sequence that have fewer than $M$ past neighbors, the summations above are taken only over valid indices with $i-p \ge 1$, and the remaining terms are omitted. This approach not only leverages historical context but also allows for dynamic adjustments based on specific temporal relationships, thereby enhancing the model's capacity to learn temporal dependencies.

\textbf{Channel Aggregation Module.} The channel aggregation process is consistent with the method outlined in \equref{eq:ffn}, which models the dependencies across different channels. Hence, we obtain the overall output with the stacked $L$ GLFormer layers, represented as $\bm{H}_{\mathrm{TA}}^{(L)} \in \mathbb{R}^{N\times d}$. We employ a mean-aggregated operator to generate the temporal representations of nodes, which is calculated as follows:
\begin{equation}
    \bm{Z}_u = \frac{1}{N}\sum^{N}_{i=1} \bm{H}^{(L)}_{\mathrm{TA},{i, :}}. 
\end{equation}

\subsection{Hierarchical Aggregation Mechanism}
In addition to learning local information from neighboring sequences, long-range temporal dependencies are essential for grasping evolving patterns in dynamic graphs. Drawing inspiration from dilated causal convolution \cite{DBLP:journals/corr/YuK15}, we progressively select various positions and the number of neighbors for aggregation as the layer deepens, thereby expanding the receptive field. Concretely, let the layer-wise offsets be defined by
$\mathcal{R}_l \;=\; \{\, p \in \mathbb{Z} \mid s^{l-1} \le p \le s^{l}\,\}$. The kernel size, $K_l = s_l - s_{l-1} + 1$ also determines the number $M$ of most recent neighbors considered by the token mixer at layer $l$ (i.e., $M = K_l$). The aggregation process is defined as follows:
\begin{equation}
    \bm{H}_{i,:}^{(l)} \;=\; \sum_{p \in \mathcal{R}_l} \big(\alpha_{p}^{i}\big)^{(l)}\,\bm{H}^{(l-1)}_{\mathrm{TA},\, i-p \,:}\,,
\end{equation}
where $\bm{H}_{i,:}^{(l)}$ denotes the aggregated representation for neighbor $u_i$ at the $l$-th layer's token mixer, and $\big(\alpha_{p}^{i}\big)^{(l)}$ are scores that modulate the influence of previous time steps $\bm{H}^{(l-1)}_{\mathrm{TA},\, i-p \,:}$. Causally invalid indices ($i-p<1$) are masked before computing the scores. This formulation aggregates contributions from a contiguous range of past time steps and enlarges the receptive field as $l$ increases.

The hierarchical structure effectively captures both short-term dynamics and long-term dependencies within the data. By utilizing multiple layers, we empower the model to learn complex relationships across various temporal scales, thereby enhancing its ability to generalize across different sequences and improve performance on temporal tasks.

\subsection{Model Training Process}
Given the interaction $e=(u_i, v_j, t)$, we obtain the probability of predicted interaction via an MLP operator based on their temporal representations $\bm{Z}_{u_i}$ and $\bm{Z}_{v_j}$. The process is:
\begin{equation}
    z_{u_i, v_j, t} = \bm{K}_2\!\Big(\mathrm{ReLU}\!\big(\bm{K}_1([\,\bm{Z}_{u_i}; \bm{Z}_{v_j}\,])\big)\Big),
\end{equation}
where $\bm{K}_1,\bm{K}_2$ are learnable parameters and $\mathrm{ReLU}(\cdot)$ is the activation function. 

We approach dynamic link prediction as a temporal binary classification problem, where the presence or absence of a link at time $t$ is the label. For an interaction $(u_i,v_j,t)$, the ground truth is $y_{u_i,v_j,t}\in\{0,1\}$, with $y_{u_i,v_j,t}=1$ indicating that a link exists between $u_i$ and $v_j$ at time $t$. We adopt a 1:1 negative sampling strategy: for each positive $(u_i,v_j,t)\in\mathcal{G}^{+}$, we sample $v_k\sim P_n(\mathcal{V}\setminus\{v_j\})$ and form a negative $(u_i,v_k,t)\in\mathcal{G}^{-}$. Let $z_{u_i,v_j,t}$ be the score output by the link predictor and $\hat{y}_{u_i,v_j,t}=\sigma(z_{u_i,v_j,t})$ the predicted probability (with $\sigma$ the sigmoid function). We train by minimizing binary cross-entropy on probabilities:
\begin{equation}
\label{equ:optimization}
\mathcal{L}
= -\sum_{(u_i,v_j,t)\in\mathcal{G}^{+}} \log \hat{y}_{u_i,v_j,t}
  -\sum_{(u_i,v_k,t)\in\mathcal{G}^{-}} \log \big(1-\hat{y}_{u_i,v_k,t}\big).
\end{equation}


\subsection{Complexity Analysis}
Let $N$ be the sequence length and $K_l=|\mathcal{R}_l|=s^{l}-s^{l-1}+1$ the layer-$l$ kernel size.
Our mixer computes one score $\big(\alpha_{p}^{i}\big)^{(l)}$ per offset $p\in\mathcal{R}_l$ and one score-weighted sum per position $i$, without pairwise interactions within the window.
Hence the per-layer time complexity is
$O(NK_l d)\ \approx\ O(NK_l)$, the number of kernel parameters per layer is $O(K_l)$ for $\{\bm{w}^{(l)}_{p}\}$.
Stacking $L$ layers yields total time $O\!\big(\sum_{l=1}^{L} N K_l\big)$.
This is markedly cheaper than full self-attention $O(N^2)$ while expanding the receptive field by enlarging the gapped interval $\mathcal{R}_l$ across layers.

\section{Experiments}
We conduct experiments on six benchmarks to evaluate the effectiveness and efficiency of our approaches.
\label{section-4}
\subsection{Datasets and Baselines}
\textbf{Datasets.} This study employs six publicly available datasets, each providing unique insights into user interactions and behaviors: Wikipedia, Reddit, MOOC, LastFM, SocialEvo, and Enron, collected in \cite{DBLP:conf/nips/PoursafaeiHPR22}. 

\textbf{Backbones.} We apply our method to several recent continuous-time dynamic graph learning models. These backbones span various techniques, including memory networks (i.e., TGN \cite{DBLP:journals/corr/abs-2006-10637}), graph convolutions (i.e., TGAT \cite{DBLP:conf/iclr/XuRKKA20}), random walks (i.e., CAWN \cite{DBLP:conf/iclr/WangCLL021}), and sequential models (i.e., TCL \cite{DBLP:journals/corr/abs-2105-07944}, and DyGFormer \cite{DBLP:conf/nips/0004S0L23}). 

\textbf{Baselines.} For comparison, we replace the Transformer's self-attention with either a pooling or Multi-Layer Perceptron (MLP) module, keeping the rest of the architecture consistent with the vanilla version. The token mixer input is denoted as $\bm{I}_u = [\bm{X}_{u_1}, \ldots, \bm{X}_{u_N}] \in \mathbb{R}^{N \times d}$, where $\bm{X}_{u_i}$ is the embedding of node $u$'s $i$-th neighbor, $N$ is the number of neighbors, and $d$ is the embedding dimension. The aggregation methods are as follows:
 
\begin{itemize}
    \item Pooling. We apply an average pooling operation over the most recent $s$ neighbors to generate fused representations for each node, effectively capturing both temporal and structural information. The number of selected neighbors is consistent with the configuration used in our model. The pooling operation is defined as:
    \begin{equation}
        \bm{H}_{u_j} = \frac{1}{s} \sum_{i=j-s+1}^{j} \bm{X}_{u_i}.
\end{equation}

    \item MLP. Alternatively, the embeddings of the neighbors are passed through a Multi-Layer Perceptron to allow for non-linear transformations:
    \begin{equation}
\bm{H}_u = \bm{W}_2 \cdot \sigma(\bm{W}_1 \cdot \bm{I}_u + \bm{b}_1) + \bm{b}_2,
    \end{equation}
    where $\bm{W}_1 \in \mathbb{R}^{\gamma N \times N}$ and $\bm{W}_2 \in \mathbb{R}^{N \times \gamma N}$ are weight matrices, $\bm{b}_1$ and $\bm{b}_2$ are biases, and $\sigma$ is the activation function (e.g., ReLU or GELU). We set the scaling factor $\gamma$ to 0.5 for all methods. This approach enables the model to learn complex interactions between neighbors.
\end{itemize} 

\begin{table*}[ht]
\resizebox{\textwidth}{!}
{
\setlength{}{}
{
\begin{tabular}{c|c|c|cccccc|c}
\hline
\multirow{2}{*}{Metric} & \multirow{2}{*}{Backbone}  & \multirow{2}{*}{Method} & \multicolumn{6}{c|}{Datasets}                                                                                                                                                                                                                                      & \multirow{2}{*}{Rank} \\ \cline{4-9}
                        &                            &                         & \multicolumn{1}{c|}{Wikipedia}               & \multicolumn{1}{c|}{Reddit}                  & \multicolumn{1}{c|}{MOOC}                    & \multicolumn{1}{c|}{LastFM}                  & \multicolumn{1}{c|}{SocialEvo}               & Enron                   &                       \\ \hline
\multirow{20}{*}{AP}    & \multirow{4}{*}{TGN}       & Vanilla                   & \multicolumn{1}{c|}{98.45 ± 0.02}            & \multicolumn{1}{c|}{98.63 ± 0.06}            & \multicolumn{1}{c|}{89.15 ± 1.60}            & \multicolumn{1}{c|}{77.07 ± 3.97}            & \multicolumn{1}{c|}{93.57 ± 0.17}            & 87.50 ± 0.90            & 3.17                  \\
                        &                            & Pooling                    & \multicolumn{1}{c|}{98.27 ±   0.14}          & \multicolumn{1}{c|}{98.64 ±   0.04}          & \multicolumn{1}{c|}{\textbf{91.86 ±   0.48}} & \multicolumn{1}{c|}{85.54 ±   1.77}          & \multicolumn{1}{c|}{93.47 ± 0.33}            & 85.90 ± 1.60            & 2.83                  \\
                        &                            & MLP                     & \multicolumn{1}{c|}{98.52 ±   0.09}          & \multicolumn{1}{c|}{98.60 ±   0.09}          & \multicolumn{1}{c|}{\textbf{91.86 ±   0.62}} & \multicolumn{1}{c|}{\textbf{86.38 ±   1.25}} & \multicolumn{1}{c|}{93.70 ± 0.00}            & 87.46 ± 1.01            & 2.17                  \\
                        &                            & GLFormer              & \multicolumn{1}{c|}{\textbf{98.71 ± 0.02}}   & \multicolumn{1}{c|}{\textbf{98.71 ± 0.02}}   & \multicolumn{1}{c|}{91.14 ± 0.56}            & \multicolumn{1}{c|}{84.96 ± 1.94}   & \multicolumn{1}{c|}{\textbf{93.91 ±   0.10}} & \textbf{88.48 ±   0.48} & \textbf{1.67}         \\ \cline{2-10} 
                        & \multirow{4}{*}{TCL}       & Vanilla                   & \multicolumn{1}{c|}{96.47 ± 0.16}            & \multicolumn{1}{c|}{97.53 ± 0.02}            & \multicolumn{1}{c|}{82.38 ± 0.24}            & \multicolumn{1}{c|}{74.75 ± 3.77}   & \multicolumn{1}{c|}{93.13 ± 0.16}            & 79.70 ±   0.71          & 3.33                  \\
                        &                            & Pooling                    & \multicolumn{1}{c|}{95.47 ±   0.12}          & \multicolumn{1}{c|}{97.78 ± 0.02}            & \multicolumn{1}{c|}{81.00 ±   0.29}          & \multicolumn{1}{c|}{71.45 ± 0.53}            & \multicolumn{1}{c|}{93.33± 0.17}             & 85.04 ±   0.44          & 3.5                   \\
                        &                            & MLP                     & \multicolumn{1}{c|}{97.69 ±   0.06}          & \multicolumn{1}{c|}{\textbf{98.67 ± 0.06}}   & \multicolumn{1}{c|}{84.15 ±   1.19}          & \multicolumn{1}{c|}{74.16 ± 5.03}            & \multicolumn{1}{c|}{93.48 ± 0.21}            & 85.57 ±   0.29          & 2                     \\
                        &                            & GLFormer              & \multicolumn{1}{c|}{\textbf{97.72 ±   0.05}} & \multicolumn{1}{c|}{97.89 ± 0.05}            & \multicolumn{1}{c|}{\textbf{84.29 ±   1.46}} & \multicolumn{1}{c|}{\textbf{76.66 ±   2.22}} & \multicolumn{1}{c|}{\textbf{93.68 ± 0.16}}   & \textbf{85.88 ±   0.18} & \textbf{1.17}         \\ \cline{2-10} 
                        & \multirow{4}{*}{TGAT}      & Vanilla                   & \multicolumn{1}{c|}{96.94 ± 0.06}            & \multicolumn{1}{c|}{\textbf{98.52 ± 0.02}}   & \multicolumn{1}{c|}{85.84 ± 0.15}            & \multicolumn{1}{c|}{73.42 ±   0.21}          & \multicolumn{1}{c|}{93.16 ± 0.17}            & 71.12 ±   0.97          & 3                     \\
                        &                            & Pooling                    & \multicolumn{1}{c|}{96.91 ±   0.08}          & \multicolumn{1}{c|}{98.31 ± 0.05}            & \multicolumn{1}{c|}{86.34 ±   0.19}          & \multicolumn{1}{c|}{73.85 ±   0.03}          & \multicolumn{1}{c|}{92.98 ± 0.07}            & 71.07 ±   0.66          & 3.5                   \\
                        &                            & MLP                     & \multicolumn{1}{c|}{97.01 ±   0.11}          & \multicolumn{1}{c|}{98.39 ± 0.02}            & \multicolumn{1}{c|}{86.34 ±   0.31}          & \multicolumn{1}{c|}{74.12 ±   0.16}          & \multicolumn{1}{c|}{93.20 ± 0.14}            & 72.53 ±   0.17          & 2.17                  \\
                        &                            & GLFormer              & \multicolumn{1}{c|}{\textbf{97.05 ±   0.10}} & \multicolumn{1}{c|}{98.42 ± 0.02}            & \multicolumn{1}{c|}{\textbf{86.61 ±   0.16}} & \multicolumn{1}{c|}{\textbf{74.24 ± 0.46}}   & \multicolumn{1}{c|}{\textbf{93.24 ±   0.07}} & \textbf{75.48 ±   0.55} & \textbf{1.17}         \\ \cline{2-10} 
                        & \multirow{4}{*}{CAWN}      & Vanilla                   & \multicolumn{1}{c|}{98.76 ± 0.03}            & \multicolumn{1}{c|}{\textbf{99.11 ± 0.01}}   & \multicolumn{1}{c|}{80.15 ± 0.25}            & \multicolumn{1}{c|}{86.99 ± 0.06}            & \multicolumn{1}{c|}{84.96 ± 0.09}            & 89.56 ±   0.09          & 2.83                  \\
                        &                            & Pooling                    & \multicolumn{1}{c|}{98.68 ±   0.02}          & \multicolumn{1}{c|}{99.08 ± 0.01}            & \multicolumn{1}{c|}{82.65 ±   0.15}          & \multicolumn{1}{c|}{87.44 ± 0.13}            & \multicolumn{1}{c|}{84.88 ± 0.42}            & 90.08 ±   0.09          & 3                     \\
                        &                            & MLP                     & \multicolumn{1}{c|}{\textbf{98.80 ±   0.01}} & \multicolumn{1}{c|}{99.10 ± 0.01}            & \multicolumn{1}{c|}{\textbf{82.86 ±   0.21}} & \multicolumn{1}{c|}{87.58 ± 0.04}            & \multicolumn{1}{c|} {\textbf{85.26 ± 0.17}}            & \textbf{90.20 ±   0.02} & \textbf{1.33}         \\
                        &                            & GLFormer              & \multicolumn{1}{c|}{98.67 ±   0.07}          & \multicolumn{1}{c|}{99.10 ± 0.00}            & \multicolumn{1}{c|}{81.37 ± 0.01}            & \multicolumn{1}{c|}{\textbf{87.72 ±   0.32}} & \multicolumn{1}{c|}{84.96 ±   0.04} & 89.58 ±   0.16          & 2.5                   \\ \cline{2-10} 
                        & \multirow{4}{*}{DyGFormer} & Vanilla                   & \multicolumn{1}{c|}{\textbf{99.03 ± 0.02}}   & \multicolumn{1}{c|}{99.22 ± 0.01}            & \multicolumn{1}{c|}{87.52 ± 0.49}            & \multicolumn{1}{c|}{93.00 ± 0.12}            & \multicolumn{1}{c|}{94.73 ± 0.01}            & 92.47 ± 0.12            & 2.17                  \\
                        &                            & Pooling                    & \multicolumn{1}{c|}{99.00 ±   0.01}          & \multicolumn{1}{c|}{99.04 ±   0.02}          & \multicolumn{1}{c|}{84.91 ±   0.39}          & \multicolumn{1}{c|}{93.20 ±   0.03}          & \multicolumn{1}{c|}{94.71 ± 0.03}            & 92.38 ± 0.11            & 3.17                  \\
                        &                            & MLP                     & \multicolumn{1}{c|}{99.00 ±   0.03}          & \multicolumn{1}{c|}{99.01 ±   0.06}          & \multicolumn{1}{c|}{85.27 ±   0.30}          & \multicolumn{1}{c|}{93.30 ±   0.04}          & \multicolumn{1}{c|}{94.66 ± 0.12}            & 92.36 ± 0.10            & 3.33                  \\
                        &                            & GLFormer              & \multicolumn{1}{c|}{\textbf{99.03 ±   0.01}} & \multicolumn{1}{c|}{\textbf{99.24 ± 0.00}}   & \multicolumn{1}{c|}{\textbf{87.87 ± 0.50}}   & \multicolumn{1}{c|}{\textbf{93.34 ± 0.16}}   & \multicolumn{1}{c|}{\textbf{94.76 ± 0.01}}   & \textbf{92.62 ± 0.06}   & \textbf{1}            \\ \hline
\multirow{20}{*}{AUC-ROC}   & \multirow{4}{*}{TGN}       & Vanilla                   & \multicolumn{1}{c|}{98.37 ±   0.02}          & \multicolumn{1}{c|}{98.60 ± 0.06}            & \multicolumn{1}{c|}{91.21 ± 1.15}            & \multicolumn{1}{c|}{78.47 ±   2.94}          & \multicolumn{1}{c|}{95.39 ± 0.17}            & 89.40 ± 0.81            & 3.17                  \\
                        &                            & Pooling                    & \multicolumn{1}{c|}{98.19 ± 0.14}            & \multicolumn{1}{c|}{98.61 ±   0.04}          & \multicolumn{1}{c|}{\textbf{93.50 ±   0.41}} & \multicolumn{1}{c|}{85.67 ± 1.42}            & \multicolumn{1}{c|}{95.24 ± 0.35}            & 87.74 ±   1.68          & 2.83                  \\
                        &                            & MLP                     & \multicolumn{1}{c|}{98.44 ± 0.09}            & \multicolumn{1}{c|}{98.24 ±   0.14}          & \multicolumn{1}{c|}{93.36 ±   0.60}          & \multicolumn{1}{c|}{\textbf{86.23 ± 1.30}}   & \multicolumn{1}{c|}{95.59 ± 0.01}            & 89.05 ±   0.78          & 2.33                  \\
                        &                            & GLFormer              & \multicolumn{1}{c|}{\textbf{98.66 ±   0.02}} & \multicolumn{1}{c|}{\textbf{98.66 ± 0.02}}   & \multicolumn{1}{c|}{92.58 ± 0.31}            & \multicolumn{1}{c|}{{85.18 ± 1.78}}   & \multicolumn{1}{c|}{\textbf{95.70 ± 0.10}}   & \textbf{90.04 ± 0.69}   & \textbf{1.67}         \\ \cline{2-10} 
                        & \multirow{4}{*}{TCL}       & Vanilla                   & \multicolumn{1}{c|}{95.84 ± 0.18}            & \multicolumn{1}{c|}{97.42 ±   0.02}          & \multicolumn{1}{c|}{83.12 ±   0.18}          & \multicolumn{1}{c|}{{69.58 ± 3.96}}   & \multicolumn{1}{c|}{94.84 ± 0.17}            & 75.74 ±   0.72          & 3.33                  \\
                        &                            & Pooling                    & \multicolumn{1}{c|}{94.97 ± 0.14}            & \multicolumn{1}{c|}{97.70 ± 0.02}            & \multicolumn{1}{c|}{82.13 ±   0.28}          & \multicolumn{1}{c|}{66.08 ± 0.66}            & \multicolumn{1}{c|}{95.18 ± 0.24}            & 83.26 ± 0.78            & 3.5                   \\
                        &                            & MLP                     & \multicolumn{1}{c|}{97.20 ± 0.09}            & \multicolumn{1}{c|}{\textbf{98.65 ± 0.06}}   & \multicolumn{1}{c|}{84.25 ±   0.79}          & \multicolumn{1}{c|}{69.16 ± 3.63}            & \multicolumn{1}{c|}{\textbf{95.62 ± 0.15}}   & 83.55 ± 0.64            & 1.83                  \\
                        &                            & GLFormer              & \multicolumn{1}{c|}{\textbf{97.27 ± 0.09}}   & \multicolumn{1}{c|}{97.79 ± 0.04}            & \multicolumn{1}{c|}{\textbf{84.40 ± 1.04}}   & \multicolumn{1}{c|}{\textbf{71.05 ± 2.21}}   & \multicolumn{1}{c|}{{95.31 ± 0.22}}   & \textbf{84.00 ± 0.26}   & \textbf{1.33}         \\ \cline{2-10} 
                        & \multirow{4}{*}{TGAT}      & Vanilla                   & \multicolumn{1}{c|}{96.67 ± 0.07}            & \multicolumn{1}{c|}{\textbf{98.47 ± 0.02}}   & \multicolumn{1}{c|}{87.11 ± 0.19}            & \multicolumn{1}{c|}{71.59 ± 0.18}            & \multicolumn{1}{c|}{94.76 ± 0.16}            & 68.89 ± 1.10            & 3.33                  \\
                        &                            & Pooling                    & \multicolumn{1}{c|}{96.70 ± 0.09}            & \multicolumn{1}{c|}{98.25 ± 0.07}            & \multicolumn{1}{c|}{87.77 ±   0.22}          & \multicolumn{1}{c|}{71.85 ± 0.12}            & \multicolumn{1}{c|}{94.55 ± 0.11}            & 69.57 ± 0.69            & 3.17                  \\
                        &                            & MLP                     & \multicolumn{1}{c|}{96.76 ± 0.12}            & \multicolumn{1}{c|}{98.32 ± 0.03}            & \multicolumn{1}{c|}{87.75 ±   0.30}          & \multicolumn{1}{c|}{72.17 ± 0.04}            & \multicolumn{1}{c|}{94.89 ± 0.09}            & 69.92 ± 0.21            & 2.33                  \\
                        &                            & GLFormer              & \multicolumn{1}{c|}{\textbf{96.81 ± 0.11}}   & \multicolumn{1}{c|}{98.34 ± 0.02}            & \multicolumn{1}{c|}{\textbf{87.81 ±   0.21}} & \multicolumn{1}{c|}{\textbf{72.39 ± 0.21}}   & \multicolumn{1}{c|}{\textbf{94.94 ± 0.05}}   & \textbf{72.56 ± 0.99}   & \textbf{1.17}         \\ \cline{2-10} 
                        & \multirow{4}{*}{CAWN}      & Vanilla                   & \multicolumn{1}{c|}{98.54 ± 0.04}            & \multicolumn{1}{c|}{\textbf{99.01 ±   0.01}} & \multicolumn{1}{c|}{80.38 ±   0.26}          & \multicolumn{1}{c|}{85.92 ± 0.10}            & \multicolumn{1}{c|}{87.34 ± 0.08}            & 90.45 ±   0.14          & 2.83                  \\
                        &                            & Pooling                    & \multicolumn{1}{c|}{98.42 ± 0.03}            & \multicolumn{1}{c|}{98.98 ± 0.01}            & \multicolumn{1}{c|}{83.03 ±   0.09}          & \multicolumn{1}{c|}{86.29 ± 0.12}            & \multicolumn{1}{c|}{87.21 ± 0.56}            & 90.74 ± 0.07            & 3.17                  \\
                        &                            & MLP                     & \multicolumn{1}{c|}{\textbf{98.61 ± 0.03}}   & \multicolumn{1}{c|}{\textbf{99.01 ± 0.01}}   & \multicolumn{1}{c|}{\textbf{83.13 ±   0.31}} & \multicolumn{1}{c|}{86.44 ± 0.05}            & \multicolumn{1}{c|}{\textbf{87.55 ± 0.13}}   & \textbf{90.83 ± 0.02}   & \textbf{1.17}         \\
                        &                            & GLFormer              & \multicolumn{1}{c|}{98.44 ± 0.14}            & \multicolumn{1}{c|}{99.00 ± 0.01}            & \multicolumn{1}{c|}{81.60 ± 0.02}            & \multicolumn{1}{c|}{\textbf{86.76 ± 0.22}}   & \multicolumn{1}{c|}{{87.37 ± 0.03}}   & 90.45 ±   0.14          & 2.5                   \\ \cline{2-10} 
                        & \multirow{4}{*}{DyGFormer} & Vanilla                   & \multicolumn{1}{c|}{\textbf{98.91 ± 0.02}}   & \multicolumn{1}{c|}{99.15 ±   0.01}          & \multicolumn{1}{c|}{87.91 ± 0.58}            & \multicolumn{1}{c|}{93.05 ± 0.10}            & \multicolumn{1}{c|}{96.30 ± 0.01}            & 93.33 ±   0.13          & 2.17                  \\
                        &                            & Pooling                    & \multicolumn{1}{c|}{98.89 ± 0.01}            & \multicolumn{1}{c|}{98.94 ±   0.07}          & \multicolumn{1}{c|}{86.22 ±   0.39}          & \multicolumn{1}{c|}{93.11 ± 0.03}            & \multicolumn{1}{c|}{96.28 ± 0.04}            & 93.33 ±   0.07          & 2.83                  \\
                        &                            & MLP                     & \multicolumn{1}{c|}{98.88 ± 0.03}            & \multicolumn{1}{c|}{98.88 ±   0.05}          & \multicolumn{1}{c|}{86.11 ±   0.39}          & \multicolumn{1}{c|}{93.21 ± 0.03}            & \multicolumn{1}{c|}{96.26 ± 0.10}            & 93.15 ±   0.14          & 3.67                  \\
                        &                            & GLFormer              & \multicolumn{1}{c|}{\textbf{98.91 ± 0.02}}   & \multicolumn{1}{c|}{\textbf{99.18 ± 0.01}}   & \multicolumn{1}{c|}{\textbf{88.73 ± 0.64}}   & \multicolumn{1}{c|}{\textbf{93.31 ± 0.13}}   & \multicolumn{1}{c|}{\textbf{96.37 ± 0.02}}   & \textbf{93.47 ± 0.15}   & \textbf{1}            \\ \hline
\end{tabular}
}}
\caption{Performance for transductive dynamic link prediction on datasets.}
\label{transductive}
\end{table*}

\subsection{Experimental Settings}
\textbf{Evaluation Tasks and Metrics.} In this work, we focus on the dynamic link prediction task, following prior works \cite{DBLP:conf/iclr/XuRKKA20,DBLP:journals/corr/abs-2006-10637,DBLP:conf/iclr/WangCLL021,DBLP:conf/nips/PoursafaeiHPR22}. We evaluate the performance in the transductive setting, where the goal is to predict future interactions between nodes that were observed during training. During training, for each positive pair, we sample a negative pair by keeping the same source node and selecting a random destination node. We assess performance using Average Precision (AP) and Area Under the Receiver Operating Characteristic Curve (AUC-ROC) as evaluation metrics. The datasets are chronologically split into 70\% for training, 15\% for validation, and 15\% for testing.

\textbf{Implementation Details.} 
To ensure fair performance comparisons, we follow the settings as described in \cite{DBLP:conf/nips/0004S0L23}. We train the models using the Adam optimizer over 100 epochs, applying early stopping after 20 epochs without improvement. The best-performing model on the validation set is selected for testing. For all datasets, we set the learning rate to 0.0001 and the batch size to 200. We search for the optimal number of GLFormer layers in the range $[1, 2, 3]$. The number of neighbors aggregated in each GLFormer layer for different models is the same as in \citet{DBLP:conf/nips/0004S0L23}. Across all models, the dimensions for both node and edge features are set to 172, while the time encoding dimension is fixed at 100. All other settings remain consistent with those specified in the original papers. Experiments are conducted on an Ubuntu machine equipped with an Intel(R) Xeon(R) Gold 6130 CPU @ 2.10GHz (16 physical cores) and an NVIDIA Tesla T4 GPU with 15 GB of memory. Our code is available at \url{https://github.com/Hope-Rita/GLFormer}.

\subsection{Performance Comparison and Discussions}
We present the performance of different methods on the AP and AUC metrics for dynamic link prediction in \tabref{transductive}. To enhance readability, the results have been multiplied by 100. The best results are highlighted in bold. Based on the results, we make the following key observations:

(i) Compared with the vanilla Transformer, which relies on self-attention and spatial smoothing over sequences and thus often mixes in irrelevant information, our approach preserves the key information at each node while selectively fusing information only from relevant positions. This targeted fusion avoids over-smoothing, ensuring that the model captures both local and essential contextual information, resulting in efficient dynamic graph learning with reduced computational overhead and redundancy compared to the vanilla Transformer.

(ii) Compared with traditional pooling methods, which aggregate features from neighboring nodes to summarize local information but often overlook temporal significance and fail to capture long-range temporal dependencies, our approach adaptively aggregates neighbor information based on their order and timing within a hierarchical structure. This enables us to effectively learn both short-term and long-term temporal patterns.

(iii) MLP mechanisms, which excel in modeling complex relationships through non-linear transformations, rely heavily on the depth of the network and the number of tokens to achieve the desired performance. Hence, their effectiveness can be constrained by the architecture's reliance on token quantity. In contrast, our approach is not constrained by the number of tokens, enabling us to flexibly focus on the underlying temporal relationships between nodes. This flexibility contributes to improved performance without introducing excessive complexity.

\subsection{Effectiveness of the Architecture}
In this study, we introduce the simplified Transformer architecture, GLFormer, which has an adaptive token mixer and hierarchical structure to learn both short and long-term temporal dependencies. We evaluate its effectiveness in capturing temporal interaction patterns through two key experiments: one focused on long-term information learning, and the other on analyzing model complexity.

\begin{figure}[!htbp]
    \centering
\includegraphics[width=1.0\columnwidth]{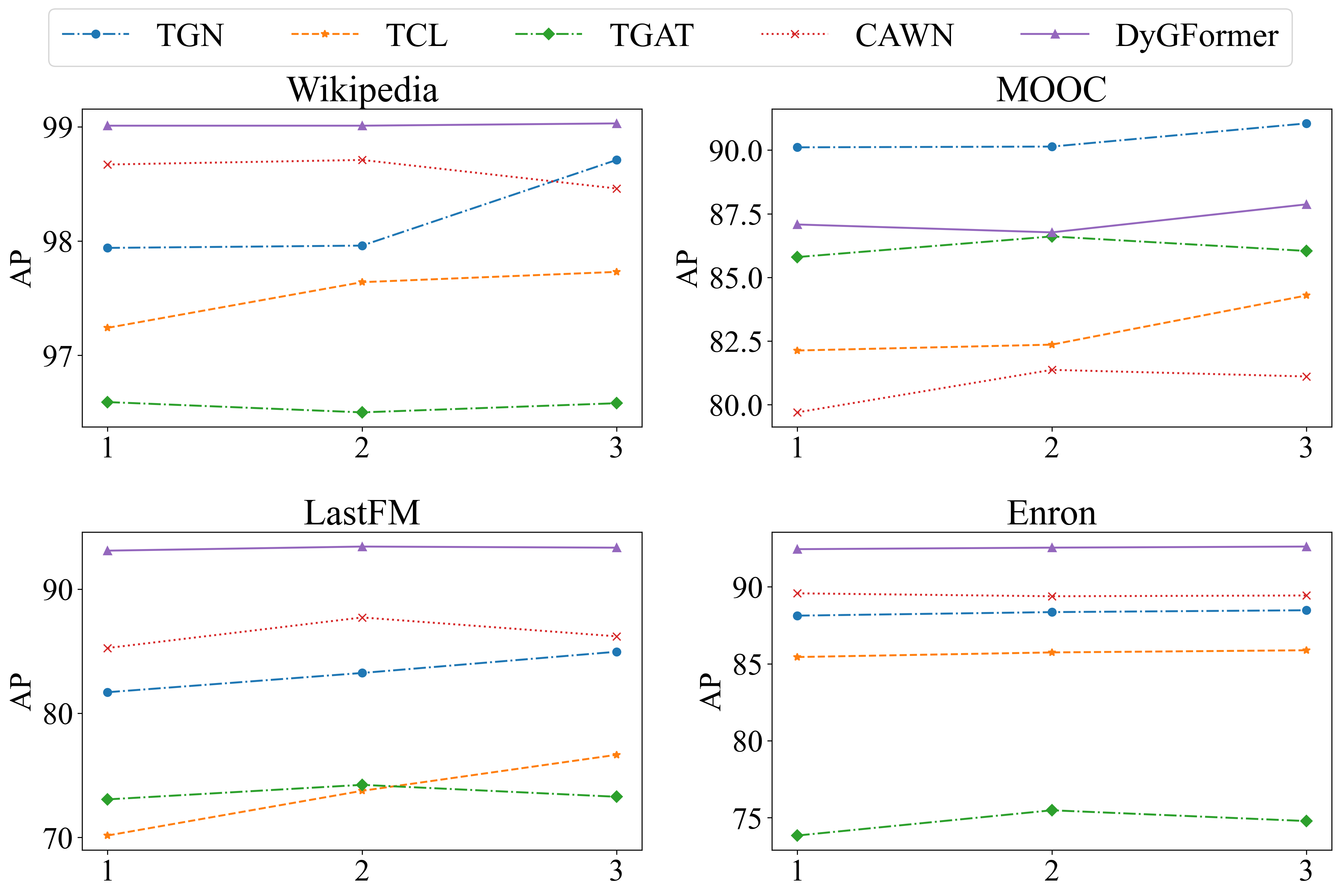}
    \caption{Performance of different numbers of GLFormer layers in transductive dynamic link prediction.}
    \label{fig:neighbors std}
\end{figure}

\textbf{Analysis of Temporal Relationship Learning}. In this study, we employ a hierarchical aggregation mechanism to capture long-term temporal dependencies. To further explore how the receptive fields of historical neighbors influence temporal information learning, we conduct experiments with varying layers, using $s^1, s^2, s^3$ set to 2, 4, and 8 for transductive link prediction tasks. The results, in terms of average precision, are presented in \figref{fig:neighbors std}.

Based on the performance in \figref{fig:neighbors std}, we observe that using a two- or three-layer GLFormer is generally sufficient for capturing temporal dependencies. For models with fewer neighbors for aggregation, such as TGN, performance shows slight variation across different layer depths. For TGAT, which employs a multi-layer graph attention mechanism for neighbors from multiple orders, a two-layer GLFormer is adequate for extracting meaningful temporal information. On the other hand, sequence-based models such as TCL and DyGFormer, which require longer sequence inputs, achieve the best performance with a three-layer GLFormer, effectively capturing long-range dependencies.

\begin{figure}[!htbp]
    \centering
\includegraphics[width=1.0\columnwidth]{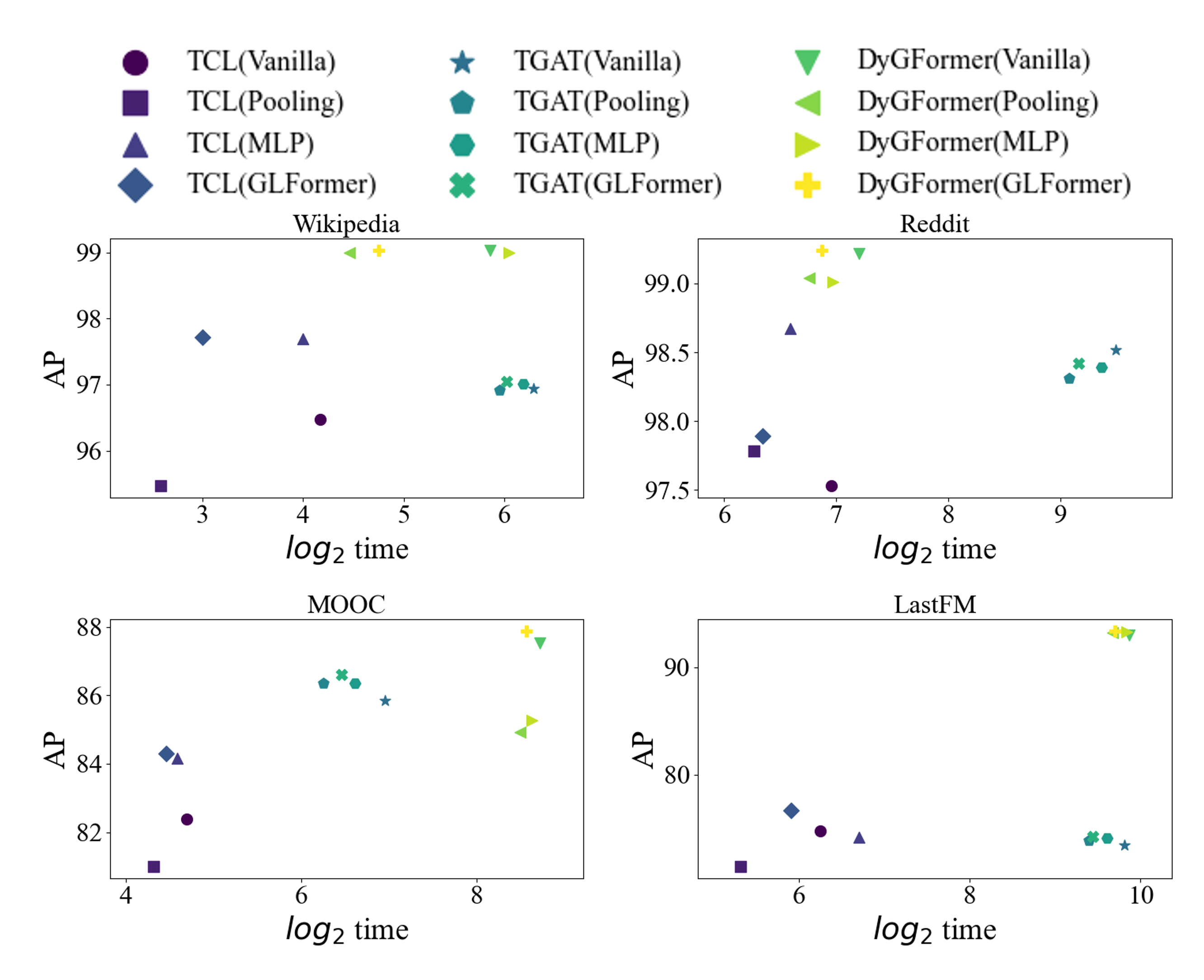}
    \caption{Log-scale evaluation time for various methods.}
    \label{fig:complexity}
\end{figure}

\textbf{Complexity Analysis}. We demonstrate the efficiency of our approach by comparing the inference time across three types of token mixers on various baseline models. Reporting inference time allows us to eliminate the influence of differing training strategies among methods. \figref{fig:complexity} compares the baselines and our GLFormer across four datasets. In contrast to the baseline models, GLFormer demonstrates a superior balance between computational efficiency and performance. This makes it particularly well-suited for real-time applications where both speed and accuracy are critical.

\subsection{Ablation Study}
\begin{figure}[!htbp]
    \centering
\includegraphics[width=1.0\columnwidth]{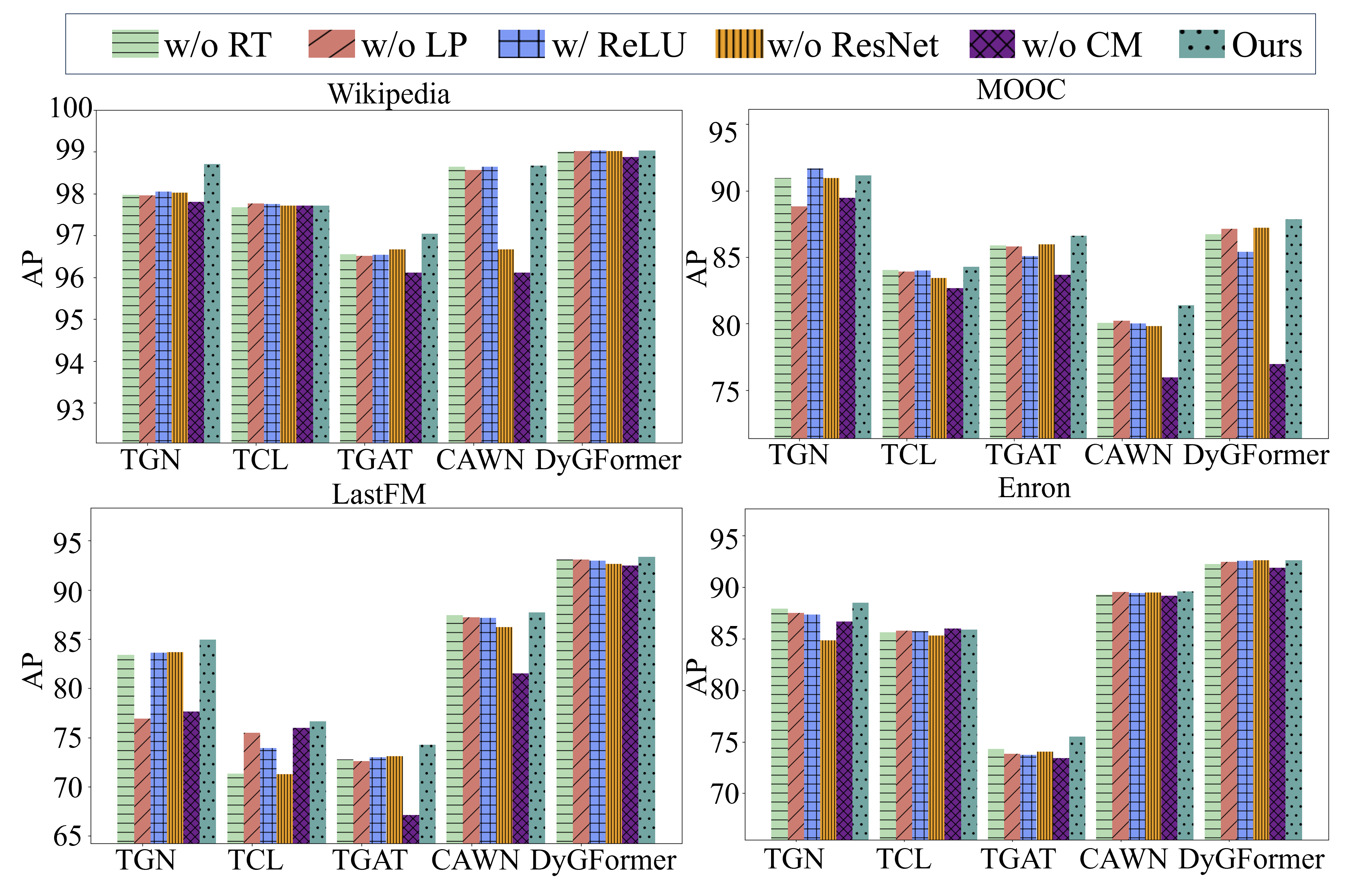}
    \caption{Effects of different components in our framework.}
    \label{fig:ablation std}
\end{figure}

We perform an ablation study to validate the effectiveness of key components within the framework. First, we assess the impact of adaptively learning the order and timing of temporal information from recent neighbors by removing the learnable parameters (LP) and the use of relative times (RT), denoted as ``w/o LP" and ``w/o RT," respectively. Additionally, we evaluate the influence of other components in the Transformer architecture. Specifically, we replace the GELU activation function with ReLU, labeled as ``w/ ReLU." We also examine the effect of removing the residual network and channel mixers, denoted as ``w/o ResNet" and ``w/o CM." The results of these experiments are illustrated in \figref{fig:ablation std}.

From the ablation study, we derive two key observations. First, both aggregation mechanisms are crucial, as they extract different but complementary information. The learnable parameters (LP) focus on capturing the temporal order and structure of interactions, while the relative times (RT) are essential for modeling the timing of events. Second, the choice of activation function, as well as the inclusion of channel mixers and the residual network, significantly impacts the framework’s performance. GELU provides smoother and more refined non-linear transformations than ReLU, resulting in better learning. The channel mixers enhance the model's ability to capture cross-channel dependencies, while the residual connections improve gradient flow and stabilize training, especially for deeper models. These components are critical for ensuring the framework's robustness and effectiveness.




\section{Conclusion}
\label{section-6}
In this work, we addressed the critical challenge of dynamic graph learning by introducing a novel Transformer framework that simplifies the traditional self-attention mechanism. We first highlighted the importance of the Transformer architecture in learning temporal dependencies among neighbors. We then proposed an adaptive token mixer that considers both temporal order and timing information simultaneously within a sliding window. To further capture the long-range temporal dependencies, we designed a hierarchical learning module, which expands the receptive fields by aggregating long-range neighbors as the layer increases. Our experimental results confirmed that this integrated approach retained the ability to model both short-term and long-term relationships effectively and demonstrated that simpler architectures can achieve competitive results. 

\section{Acknowledgements}
This work was supported by the National Natural Science Foundation of China under Grant No. U2469205, the Fundamental Research Funds for the Central Universities of China under Grant No. JKF-20240769.

\bibliography{reference}

\begin{thebibliography}{45}
\providecommand{\natexlab}[1]{#1}

\bibitem[{Chen, O'Bray, and Borgwardt(2022)}]{DBLP:conf/icml/ChenOB22}
Chen, D.; O'Bray, L.; and Borgwardt, K.~M. 2022.
\newblock Structure-Aware Transformer for Graph Representation Learning.
\newblock In \emph{{ICML} 2022, 17-23 July 2022, Baltimore, Maryland, {USA}},
  volume 162 of \emph{Proceedings of Machine Learning Research}, 3469--3489.
  {PMLR}.

\bibitem[{Chen et~al.(2024)Chen, Xiong, Zhang, Zhang, Zhang, Zhou, Wu, Zhang,
  Liu, and Wang}]{chen2024dtformer}
Chen, X.; Xiong, Y.; Zhang, S.; Zhang, J.; Zhang, Y.; Zhou, S.; Wu, X.; Zhang,
  M.; Liu, T.; and Wang, W. 2024.
\newblock Dtformer: A transformer-based method for discrete-time dynamic graph
  representation learning.
\newblock In \emph{Proceedings of the 33rd ACM International Conference on
  Information and Knowledge Management}, 301--311.

\bibitem[{Cheng et~al.(2025)Cheng, Peng, Wang, Chang, Ye, and
  Du}]{DBLP:conf/kdd/0003PWCY025}
Cheng, K.; Peng, L.; Wang, P.; Chang, H.; Ye, J.; and Du, B. 2025.
\newblock On the Scalability of Temporal Relative Positional Encoding for
  Dynamic Link Prediction.
\newblock In \emph{{KDD} 2025, Toronto ON, Canada, August 3-7, 2025}, 298--309.
  {ACM}.

\bibitem[{Cheng et~al.(2024)Cheng, Peng, Ye, Sun, and
  Du}]{DBLP:conf/kdd/0003PYS024}
Cheng, K.; Peng, L.; Ye, J.; Sun, L.; and Du, B. 2024.
\newblock Co-Neighbor Encoding Schema: {A} Light-cost Structure Encoding Method
  for Dynamic Link Prediction.
\newblock In \emph{{KDD} 2024, Barcelona, Spain, August 25-29, 2024}, 421--432.
  {ACM}.

\bibitem[{de~Barros et~al.(2023)de~Barros, Mendon{\c{c}}a, Vieira, and
  Ziviani}]{DBLP:journals/csur/BarrosMVZ23}
de~Barros, C. D.~T.; Mendon{\c{c}}a, M. R.~F.; Vieira, A.~B.; and Ziviani, A.
  2023.
\newblock A Survey on Embedding Dynamic Graphs.
\newblock \emph{{ACM} Comput. Surv.}, 55(2): 10:1--10:37.

\bibitem[{Devlin et~al.(2019)Devlin, Chang, Lee, and
  Toutanova}]{DBLP:conf/naacl/DevlinCLT19}
Devlin, J.; Chang, M.; Lee, K.; and Toutanova, K. 2019.
\newblock {BERT:} Pre-training of Deep Bidirectional Transformers for Language
  Understanding.
\newblock In \emph{{NAACL-HLT} 2019, Minneapolis, MN, USA, June 2-7, 2019,
  Volume 1 (Long and Short Papers)}, 4171--4186. Association for Computational
  Linguistics.

\bibitem[{Ding et~al.(2024)Ding, Li, He, Norelli, Wu, Tresp, Bronstein, and
  Ma}]{ding2024dygmamba}
Ding, Z.; Li, Y.; He, Y.; Norelli, A.; Wu, J.; Tresp, V.; Bronstein, M.; and
  Ma, Y. 2024.
\newblock Dygmamba: Efficiently modeling long-term temporal dependency on
  continuous-time dynamic graphs with state space models.
\newblock \emph{arXiv preprint arXiv:2408.04713}.

\bibitem[{Dosovitskiy et~al.(2021)Dosovitskiy, Beyer, Kolesnikov, Weissenborn,
  Zhai, Unterthiner, Dehghani, Minderer, Heigold, Gelly, Uszkoreit, and
  Houlsby}]{DBLP:conf/iclr/DosovitskiyB0WZ21}
Dosovitskiy, A.; Beyer, L.; Kolesnikov, A.; Weissenborn, D.; Zhai, X.;
  Unterthiner, T.; Dehghani, M.; Minderer, M.; Heigold, G.; Gelly, S.;
  Uszkoreit, J.; and Houlsby, N. 2021.
\newblock An Image is Worth 16x16 Words: Transformers for Image Recognition at
  Scale.
\newblock In \emph{9th International Conference on Learning Representations,
  {ICLR} 2021, Virtual Event, Austria, May 3-7, 2021}. OpenReview.net.

\bibitem[{Jain, Katarya, and Sachdeva(2023)}]{DBLP:journals/tweb/JainKS23}
Jain, L.; Katarya, R.; and Sachdeva, S. 2023.
\newblock Opinion Leaders for Information Diffusion Using Graph Neural Network
  in Online Social Networks.
\newblock \emph{{ACM} Trans. Web}, 17(2): 13:1--13:37.

\bibitem[{Ji et~al.(2022)Ji, Wang, Jiang, Jiang, and
  Zhang}]{DBLP:conf/aaai/JiW0JZ22}
Ji, J.; Wang, J.; Jiang, Z.; Jiang, J.; and Zhang, H. 2022.
\newblock {STDEN:} Towards Physics-Guided Neural Networks for Traffic Flow
  Prediction.
\newblock In \emph{{AAAI} 2022, February 22 - March 1, 2022}, 4048--4056.
  {AAAI} Press.

\bibitem[{Ji et~al.(2024)Ji, Liu, Sun, Liu, and Zhu}]{ji2024memmap}
Ji, S.; Liu, M.; Sun, L.; Liu, C.; and Zhu, T. 2024.
\newblock Memmap: An adaptive and latent memory structure for dynamic graph
  learning.
\newblock In \emph{Proceedings of the 30th ACM SIGKDD Conference on Knowledge
  Discovery and Data Mining}, 1257--1268.

\bibitem[{Jiang, Huang, and Huang(2023)}]{DBLP:conf/kdd/Jiang0H23}
Jiang, Y.; Huang, C.; and Huang, L. 2023.
\newblock Adaptive Graph Contrastive Learning for Recommendation.
\newblock In Singh, A.~K.; Sun, Y.; Akoglu, L.; Gunopulos, D.; Yan, X.; Kumar,
  R.; Ozcan, F.; and Ye, J., eds., \emph{{KDD} 2023, Long Beach, CA, USA,
  August 6-10, 2023}, 4252--4261. {ACM}.

\bibitem[{Jin et~al.(2024)Jin, Liang, Fang, Shao, Huang, Zhang, and
  Zheng}]{DBLP:journals/tkde/JinLFSHZZ24}
Jin, G.; Liang, Y.; Fang, Y.; Shao, Z.; Huang, J.; Zhang, J.; and Zheng, Y.
  2024.
\newblock Spatio-Temporal Graph Neural Networks for Predictive Learning in
  Urban Computing: {A} Survey.
\newblock \emph{{IEEE} Trans. Knowl. Data Eng.}, 36(10): 5388--5408.

\bibitem[{Kreuzer et~al.(2021)Kreuzer, Beaini, Hamilton, L{\'{e}}tourneau, and
  Tossou}]{DBLP:conf/nips/KreuzerBHLT21}
Kreuzer, D.; Beaini, D.; Hamilton, W.~L.; L{\'{e}}tourneau, V.; and Tossou, P.
  2021.
\newblock Rethinking Graph Transformers with Spectral Attention.
\newblock In \emph{NeurIPS 2021, December 6-14, 2021, virtual}, 21618--21629.

\bibitem[{Kumar, Zhang, and Leskovec(2019)}]{DBLP:conf/kdd/KumarZL19}
Kumar, S.; Zhang, X.; and Leskovec, J. 2019.
\newblock Predicting Dynamic Embedding Trajectory in Temporal Interaction
  Networks.
\newblock In \emph{{KDD}' 19, Anchorage, AK, USA, August 4-8, 2019},
  1269--1278. {ACM}.

\bibitem[{Liu et~al.(2022)Liu, Yu, Liao, Li, Lin, Liu, and
  Dustdar}]{DBLP:conf/iclr/LiuYLLLLD22}
Liu, S.; Yu, H.; Liao, C.; Li, J.; Lin, W.; Liu, A.~X.; and Dustdar, S. 2022.
\newblock Pyraformer: Low-Complexity Pyramidal Attention for Long-Range Time
  Series Modeling and Forecasting.
\newblock In \emph{{ICLR} 2022, April 25-29, 2022}. OpenReview.net.

\bibitem[{Liu et~al.(2021)Liu, Lin, Cao, Hu, Wei, Zhang, Lin, and
  Guo}]{DBLP:conf/iccv/LiuL00W0LG21}
Liu, Z.; Lin, Y.; Cao, Y.; Hu, H.; Wei, Y.; Zhang, Z.; Lin, S.; and Guo, B.
  2021.
\newblock Swin Transformer: Hierarchical Vision Transformer using Shifted
  Windows.
\newblock In \emph{{ICCV} 2021, Montreal, QC, Canada, October 10-17, 2021},
  9992--10002. {IEEE}.

\bibitem[{Nguyen et~al.(2018)Nguyen, Lee, Rossi, Ahmed, Koh, and
  Kim}]{DBLP:conf/bigdataconf/NguyenLRAKK18}
Nguyen, G.~H.; Lee, J.~B.; Rossi, R.~A.; Ahmed, N.~K.; Koh, E.; and Kim, S.
  2018.
\newblock Dynamic Network Embeddings: From Random Walks to Temporal Random
  Walks.
\newblock In \emph{{(IEEE} BigData 2018), Seattle, WA, USA, December 10-13,
  2018}, 1085--1092. {IEEE}.

\bibitem[{Ni et~al.(2023)Ni, Yu, Liu, Li, and Lin}]{DBLP:conf/nips/NiYLLL23}
Ni, Z.; Yu, H.; Liu, S.; Li, J.; and Lin, W. 2023.
\newblock BasisFormer: Attention-based Time Series Forecasting with Learnable
  and Interpretable Basis.
\newblock In \emph{NeurIPS 2023, New Orleans, LA, USA, December 10 - 16, 2023}.

\bibitem[{Pan et~al.(2025)Pan, Gao, Cai, Chen, and Li}]{pan2025light}
Pan, Z.; Gao, C.; Cai, F.; Chen, H.; and Li, Y. 2025.
\newblock Light Dynamic Graph Learning on Temporal Networks.
\newblock \emph{ACM Transactions on Information Systems}.

\bibitem[{Park and Kim(2022)}]{DBLP:conf/iclr/ParkK22}
Park, N.; and Kim, S. 2022.
\newblock How Do Vision Transformers Work?
\newblock In \emph{{ICLR} 2022, Virtual Event, April 25-29, 2022}.
  OpenReview.net.

\bibitem[{Patel et~al.(2023)Patel, Li, Rasooli, Constant, Raffel, and
  Callison{-}Burch}]{DBLP:conf/iclr/PatelLRCRC23}
Patel, A.; Li, B.; Rasooli, M.~S.; Constant, N.; Raffel, C.; and
  Callison{-}Burch, C. 2023.
\newblock Bidirectional Language Models Are Also Few-shot Learners.
\newblock In \emph{{ICLR} 2023, Kigali, Rwanda, May 1-5, 2023}.

\bibitem[{Poursafaei et~al.(2022)Poursafaei, Huang, Pelrine, and
  Rabbany}]{DBLP:conf/nips/PoursafaeiHPR22}
Poursafaei, F.; Huang, S.; Pelrine, K.; and Rabbany, R. 2022.
\newblock Towards Better Evaluation for Dynamic Link Prediction.
\newblock In \emph{NeurIPS 2022, New Orleans, LA, USA, November 28 - December
  9, 2022}.

\bibitem[{Rao et~al.(2022)Rao, Zhao, Tang, Zhou, Lim, and
  Lu}]{DBLP:conf/nips/RaoZT0LL22}
Rao, Y.; Zhao, W.; Tang, Y.; Zhou, J.; Lim, S.; and Lu, J. 2022.
\newblock HorNet: Efficient High-Order Spatial Interactions with Recursive
  Gated Convolutions.
\newblock In \emph{NeurIPS 2022, New Orleans, LA, USA, November 28 - December
  9, 2022}.

\bibitem[{Rossi et~al.(2020)Rossi, Chamberlain, Frasca, Eynard, Monti, and
  Bronstein}]{DBLP:journals/corr/abs-2006-10637}
Rossi, E.; Chamberlain, B.; Frasca, F.; Eynard, D.; Monti, F.; and Bronstein,
  M.~M. 2020.
\newblock Temporal Graph Networks for Deep Learning on Dynamic Graphs.
\newblock \emph{CoRR}, abs/2006.10637.

\bibitem[{Steiner et~al.(2022)Steiner, Kolesnikov, Zhai, Wightman, Uszkoreit,
  and Beyer}]{DBLP:journals/tmlr/SteinerKZWUB22}
Steiner, A.; Kolesnikov, A.; Zhai, X.; Wightman, R.; Uszkoreit, J.; and Beyer,
  L. 2022.
\newblock How to train your ViT? Data, Augmentation, and Regularization in
  Vision Transformers.
\newblock \emph{Trans. Mach. Learn. Res.}, 2022.

\bibitem[{Vaswani et~al.(2017)Vaswani, Shazeer, Parmar, Uszkoreit, Jones,
  Gomez, Kaiser, and Polosukhin}]{DBLP:conf/nips/VaswaniSPUJGKP17}
Vaswani, A.; Shazeer, N.; Parmar, N.; Uszkoreit, J.; Jones, L.; Gomez, A.~N.;
  Kaiser, L.; and Polosukhin, I. 2017.
\newblock Attention is All you Need.
\newblock In \emph{NeurIPS 2017, December 4-9, 2017, Long Beach, CA, {USA}},
  5998--6008.

\bibitem[{Wang et~al.(2021{\natexlab{a}})Wang, Chang, Li, Chu, Li, Zhang, He,
  Song, Zhou, and Yang}]{DBLP:journals/corr/abs-2105-07944}
Wang, L.; Chang, X.; Li, S.; Chu, Y.; Li, H.; Zhang, W.; He, X.; Song, L.;
  Zhou, J.; and Yang, H. 2021{\natexlab{a}}.
\newblock {TCL:} Transformer-based Dynamic Graph Modelling via Contrastive
  Learning.
\newblock \emph{CoRR}, abs/2105.07944.

\bibitem[{Wang, Chen, and Chen(2024)}]{DBLP:conf/aaai/WangCC24}
Wang, M.; Chen, W.; and Chen, B. 2024.
\newblock Considering Nonstationary within Multivariate Time Series with
  Variational Hierarchical Transformer for Forecasting.
\newblock In \emph{{AAAI} 2024, February 20-27, 2024, Vancouver, Canada},
  15563--15570. {AAAI} Press.

\bibitem[{Wang et~al.(2024)Wang, Zhou, Wen, Gao, Ding, and
  Jin}]{DBLP:conf/iclr/WangZWGD024}
Wang, X.; Zhou, T.; Wen, Q.; Gao, J.; Ding, B.; and Jin, R. 2024.
\newblock {CARD:} Channel Aligned Robust Blend Transformer for Time Series
  Forecasting.
\newblock In \emph{The Twelfth International Conference on Learning
  Representations, {ICLR} 2024, Vienna, Austria, May 7-11, 2024}.
  OpenReview.net.

\bibitem[{Wang et~al.(2021{\natexlab{b}})Wang, Chang, Liu, Leskovec, and
  Li}]{DBLP:conf/iclr/WangCLL021}
Wang, Y.; Chang, Y.; Liu, Y.; Leskovec, J.; and Li, P. 2021{\natexlab{b}}.
\newblock Inductive Representation Learning in Temporal Networks via Causal
  Anonymous Walks.
\newblock In \emph{{ICLR} 2021, Virtual Event, Austria, May 3-7, 2021}.

\bibitem[{Xu et~al.(2020)Xu, Ruan, K{\"{o}}rpeoglu, Kumar, and
  Achan}]{DBLP:conf/iclr/XuRKKA20}
Xu, D.; Ruan, C.; K{\"{o}}rpeoglu, E.; Kumar, S.; and Achan, K. 2020.
\newblock Inductive representation learning on temporal graphs.
\newblock In \emph{{ICLR} 2020, Addis Ababa, Ethiopia, April 26-30, 2020}.

\bibitem[{Xu et~al.(2025{\natexlab{a}})Xu, Zhang, Lin, and
  Zhang}]{xu2025unidyg}
Xu, Y.; Zhang, W.; Lin, X.; and Zhang, Y. 2025{\natexlab{a}}.
\newblock UniDyG: A Unified and Effective Representation Learning Approach for
  Large Dynamic Graphs.
\newblock \emph{IEEE Transactions on Knowledge and Data Engineering}.

\bibitem[{Xu et~al.(2024)Xu, Zhang, Zhang, Orlowska, and Lin}]{xu2024timesgn}
Xu, Y.; Zhang, W.; Zhang, Y.; Orlowska, M.; and Lin, X. 2024.
\newblock TimeSGN: Scalable and effective temporal graph neural network.
\newblock In \emph{2024 IEEE 40th International Conference on Data Engineering
  (ICDE)}, 3297--3310. IEEE.

\bibitem[{Xu et~al.(2025{\natexlab{b}})Xu, Zhang, Zhang, Xu, and
  Lin}]{xu2025fast}
Xu, Y.; Zhang, W.; Zhang, Y.; Xu, X.; and Lin, X. 2025{\natexlab{b}}.
\newblock Fast and accurate temporal hypergraph representation for hyperedge
  prediction.
\newblock In \emph{Proceedings of the 31st ACM SIGKDD Conference on Knowledge
  Discovery and Data Mining V. 1}, 1727--1738.

\bibitem[{Yu and Koltun(2016)}]{DBLP:journals/corr/YuK15}
Yu, F.; and Koltun, V. 2016.
\newblock Multi-Scale Context Aggregation by Dilated Convolutions.
\newblock In \emph{{ICLR} 2016, San Juan, Puerto Rico, May 2-4, 2016}.

\bibitem[{Yu et~al.(2023)Yu, Sun, Du, and Lv}]{DBLP:conf/nips/0004S0L23}
Yu, L.; Sun, L.; Du, B.; and Lv, W. 2023.
\newblock Towards Better Dynamic Graph Learning: New Architecture and Unified
  Library.
\newblock In \emph{NeurIPS 2023, New Orleans, LA, USA, December 10 - 16, 2023}.

\bibitem[{Yu et~al.(2018)Yu, Cheng, Aggarwal, Zhang, Chen, and
  Wang}]{DBLP:conf/kdd/YuCAZCW18}
Yu, W.; Cheng, W.; Aggarwal, C.~C.; Zhang, K.; Chen, H.; and Wang, W. 2018.
\newblock NetWalk: {A} Flexible Deep Embedding Approach for Anomaly Detection
  in Dynamic Networks.
\newblock In \emph{{KDD} 2018, London, UK, August 19-23, 2018}, 2672--2681.
  {ACM}.

\bibitem[{Yu et~al.(2022)Yu, Luo, Zhou, Si, Zhou, Wang, Feng, and
  Yan}]{DBLP:conf/cvpr/YuLZSZWFY22}
Yu, W.; Luo, M.; Zhou, P.; Si, C.; Zhou, Y.; Wang, X.; Feng, J.; and Yan, S.
  2022.
\newblock MetaFormer is Actually What You Need for Vision.
\newblock In \emph{{CVPR} 2022, New Orleans, LA, USA, June 18-24, 2022},
  10809--10819. {IEEE}.

\bibitem[{Yu et~al.(2024)Yu, Si, Zhou, Luo, Zhou, Feng, Yan, and
  Wang}]{DBLP:journals/pami/YuSZLZFYW24}
Yu, W.; Si, C.; Zhou, P.; Luo, M.; Zhou, Y.; Feng, J.; Yan, S.; and Wang, X.
  2024.
\newblock MetaFormer Baselines for Vision.
\newblock \emph{{IEEE} TPAMI.}, 46(2): 896--912.

\bibitem[{Yu, Liao, and Luo(2024)}]{yu2024genti}
Yu, Z.; Liao, N.; and Luo, S. 2024.
\newblock GENTI: GPU-powered Walk-based Subgraph Extraction for Scalable
  Representation Learning on Dynamic Graphs.
\newblock \emph{Proceedings of the VLDB Endowment}, 17(9): 2269--2278.

\bibitem[{Zhang et~al.(2023)Zhang, Zhang, Yan, Deng, and
  Yang}]{DBLP:journals/tois/Zhang0YD023}
Zhang, Y.; Zhang, Y.; Yan, D.; Deng, S.; and Yang, Y. 2023.
\newblock Revisiting Graph-based Recommender Systems from the Perspective of
  Variational Auto-Encoder.
\newblock \emph{{ACM} Trans. Inf. Syst.}, 41(3): 81:1--81:28.

\bibitem[{Zhou et~al.(2021)Zhou, Zhang, Peng, Zhang, Li, Xiong, and
  Zhang}]{DBLP:conf/aaai/ZhouZPZLXZ21}
Zhou, H.; Zhang, S.; Peng, J.; Zhang, S.; Li, J.; Xiong, H.; and Zhang, W.
  2021.
\newblock Informer: Beyond Efficient Transformer for Long Sequence Time-Series
  Forecasting.
\newblock In \emph{{AAAI} 2021, February 2-9, 2021}, 11106--11115. {AAAI}
  Press.

\bibitem[{Zhou et~al.(2023)Zhou, Liu, Ding, Jin, and
  Li}]{DBLP:conf/www/ZhouLDJ023}
Zhou, Z.; Liu, Y.; Ding, J.; Jin, D.; and Li, Y. 2023.
\newblock Hierarchical Knowledge Graph Learning Enabled Socioeconomic Indicator
  Prediction in Location-Based Social Network.
\newblock In \emph{{WWW} 2023, Austin, TX, USA, 30 April 2023 - 4 May 2023},
  122--132. {ACM}.

\bibitem[{Zou et~al.(2024)Zou, Mao, Ye, and Du}]{DBLP:conf/kdd/0003MY024}
Zou, T.; Mao, Y.; Ye, J.; and Du, B. 2024.
\newblock Repeat-Aware Neighbor Sampling for Dynamic Graph Learning.
\newblock In \emph{{KDD} 2024, Barcelona, Spain, August 25-29, 2024},
  4722--4733. {ACM}.

\end{thebibliography}

\end{document}